%% file: acl.tex
\pdfoutput=1

\documentclass[11pt]{article}

\usepackage{EMNLP2022}
\usepackage{times}
\usepackage{latexsym}
\usepackage{booktabs}
\usepackage{subcaption}
\usepackage{mwe}
\usepackage{comment}
\usepackage{hyperref}

\newcommand\scalemath[2]{\scalebox{#1}{\mbox{\ensuremath{\displaystyle #2}}}}
\usepackage{amsmath}

\usepackage[T1]{fontenc}
\usepackage{graphicx}
\usepackage{multirow}

\usepackage[utf8]{inputenc}

\usepackage{microtype}

\usepackage{inconsolata}

\usepackage{lipsum}

\title{Deconfounding Legal Judgment Prediction for   European Court of Human Rights Cases Towards Better Alignment with Experts}

\author{ Santosh T.Y.S.S$^{1*}$ \and Shanshan Xu$^{1*}$ \and  Oana Ichim$^2$ \and Matthias Grabmair$^1$
\\ $^1$School of Computation, Information, and Technology; Technical University of Munich, Germany \\
$^2$Graduate Institute of International and Development Studies, Geneva, Switzerland\\ 
 \texttt{\{santosh.tokala, shanshan.xu, matthias.grabmair\}@tum.de} \\
  \texttt{oana.ichim@graduateinstitute.ch}
  }

\begin{document}

\maketitle

\begin{abstract}
\input{text/abstract}
\end{abstract}

\def\thefootnote{*}\footnotetext{These authors contributed equally to this work}
\section{Introduction}
\input{text/introduction}

\section{Related Work}
\input{text/related}

\section{ECtHR Tasks \& Datasets}

\input{text/dataset}

\section{Expert-Informed Deconfounding}
\label{sec:deconfounding}
\input{text/framework}

\section{Experiments \& Discussion}
\input{text/experiments}

\section{Conclusion}
\input{text/conclusion}

\section*{Limitations}
\input{text/limitations}

\section*{Ethics Statement}
The research presented here works exclusively with publicly available datasets of ECtHR decisions, which are based on HUDOC\footnote{\url{https://hudoc.echr.coe.int}}, the public database of the Court. While these decisions are not anonymized and contain the real names of individuals involved, our work does not engage with the data in a way that we consider harmful beyond this availability.

Our models are designed to be used with pretrained language models and hence inherit any biases they may contain. This entails an obligation to screen incorporated models and to test any developed system with regard to its performance across groups of cases (e.g. \citealt{chalkidis-etal-2022-fairlex}), and to remedy any disparities before deploying it as a prediction and inference tool. Our experiments are targeted at controlling for legally irrelevant distractors in the input, which is in line with this responsibility. 

The task of legal judgment prediction raises ethical concerns, both general as well as specific to the European Court of Human Rights. \cite{fikfak2021future} emphasizes focal issues with regard to the court considering the use predictive technology to tackle its caseload, including system bias and the challenges of designing the interaction between judges and predictive systems. The latter is of course especially sensitive given experiences made with recidivism risk prediction (\citealt{collins2018punishing}) and possible disparate effects of how judges interact with scores (\citealt{albright2019if}). Our research group is committed to research on LJP as a means to derive insight from legal decision data towards increasing accountability, fairness, and transparency in the use of technology in legal systems. The premise of this work is that the behavior of legal outcome prediction systems is to be scrutinized with great care. This paper does not advocate for the practical use of such systems, but rather empirically explores difficulties that arise in their development and recommends a closer connection between technical research and legal expertise (see Sec. \ref{ljp_recommendations}). 

All models of this project were developed and trained on Google Colab. Our models adapted pretrained language models and we did not engage in any training of such large models from scratch. We did not track computation hours.

\section*{Acknowledgments}
We are grateful to Jaromir Savelka for the ability to use the \textit{Gloss} annotation tool and for providing feedback on the draft. We also thank the anonymous reviewers for valuable comments.

\bibliography{anthology,custom}
\bibliographystyle{acl_natbib}

\newpage
\appendix
\input{text/appendix}


\end{document}

%% file: text/abstract.tex
This work demonstrates that Legal Judgement Prediction systems without expert-informed adjustments can be vulnerable to shallow, distracting surface signals that arise from corpus construction, case distribution, and confounding factors. To mitigate this, we use domain expertise to strategically identify statistically predictive but legally irrelevant information. We adopt adversarial training to prevent the system from relying on it. We evaluate our deconfounded models by employing interpretability techniques and comparing to expert annotations. Quantitative experiments and qualitative analysis show that our deconfounded model consistently  aligns better with expert rationales than baselines trained for prediction only. We further contribute a set of reference expert annotations to the validation and testing partitions of an existing benchmark dataset of European Court of Human Rights cases.


%% file: text/introduction.tex
The task of Legal Judgment Prediction (LJP) has recently gained increasing attention in the legal and mainstream NLP communities \cite{aletras2016predicting, zhong-etal-2018-legal, medvedeva2020using, liu2019legal,sert2021using}. Legal cases are resolved through the exchange of arguments in front of a decision body by lawyers who represent litigating parties. This typically involves evidential reasoning, the determination of relevant rules from sources of law (e.g., codes, regulations, precedent), their application to the case, and the balancing of legal and societal values. In the NLP context, LJP takes the form of classifying the outcome of a case from some textual representation of its specific facts, effectively skipping legal reasoning. This forms a counterpoint to knowledge-focused approaches to outcome prediction (e.g., \citealp{bruninghaus2005generating,branting2013reasoning,grabmair2017predicting}) that connect to a lawyer's understanding of the domain but also require substantial knowledge engineering.

\begin{figure}[]
\includegraphics[width =0.45\textwidth]{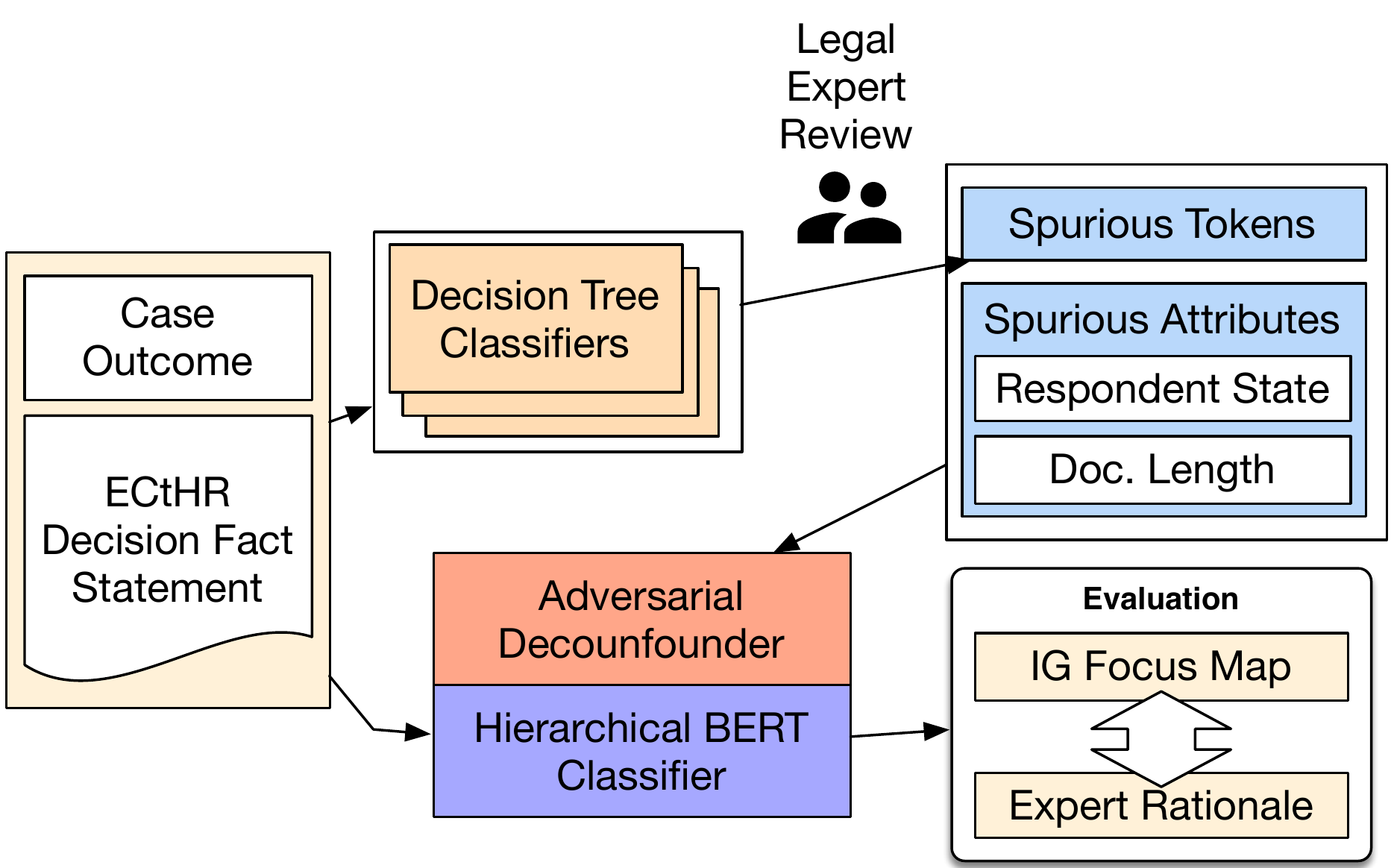}
{\caption{\label{deconfound} Our deconfounding experiment architecture}}
\end{figure}

This carries particular risk in the legal domain, where systems may rely on data elements that are statistically predictive but legally irrelevant, or even forbidden as decision criteria (e.g., the race of an accused person). This can lead to undesirable consequences, ranging from suboptimal litigation strategy decisions, flawed inference about factors predictive for the outcome, to disparate impact of decisions across groups that are to be treated equally. If legal decisions are to be informed by predictive systems processing textual case descriptions, then such systems must strive to be as closely aligned with legally relevant and permissible parts of the input as possible.

In this work, we focus on LJP for the European Court of Human Rights (ECtHR), which adjudicates complaints by individuals against states about alleged violations of their rights as enshrined in the European Convention of Human Rights. We trained deep neural models on four tasks across two existing, related datasets \cite{chalkidis-etal-2019-neural,chalkidis-etal-2022-lexglue} around predicting such violations alleged by the claimant and decided by the court. We find that the models substantially base their predictions on aspects of the text that correlate with the outcome but either have no legal bearing or are forbidden nationality-related information that stem from the distribution of cases arising at the court.


To improve the alignment of model focus with legal expert understanding, we apply a series of deconfounding measures, including a vocabulary-based method which identifies predictive tokens using a simple model. The third author, who is an ECtHR expert, then identifies distractors among them. The distracting signal can subsequently be removed from the encodings via adversarial training. This procedure is an effective way of engaging with domain experts and obtaining information about what the model should be steered away from by means of deconfounding, rather than trying to attract the model towards relevant elements via expensive data collection for supervised training. For simplicity, throughout this paper, we use `deconfounding' in an inclusive sense as the mitigation of distracting effects of (a) confounders in the statistical sense that influence both the dependent and independent variables, (b) reverse causation relationships, and (c) other attributes that spuriously correlate with the target variable. See Fig. \ref{deconfound} for an overview of our experiment design.

We evaluate our trained and deconfounded models with regard to an alignment of its explanation rationales with (1) a dataset of expert passage relevance assessments we collected and will make available to community as a supplement to \citet{chalkidis-etal-2019-neural}, and (2) on expert relevance assessments published as part of \citet{chalkidis2021paragraph}. Our results show that our deconfounding steps succeed in improving the model focus alignment with expert-identified, relevant patterns on both sets of reference annotations.

In sum, we make the following contributions:
\begin{itemize}
    \item We introduce an expert-informed deconfounding method which  identifies distracting effects from confounders and spurious correlations using a simple model, and mitigates them through adversarial training, thus helping to improve the alignment of the model focus with legal expert rationales. 
    \item We empirically evaluate this method on four tasks in legal judgment prediction on ECtHR data and show that our model consistently aligns better with expert rationales than a baseline trained for the prediction target only.
    \item We release a set of gold rationales annotated by an ECtHR  expert as a supplement to an existing dataset to facilitate future work on deriving more useful insight from trained predictive systems in the legal domain.\footnote{Our rationales and code are available at  \href{https://github.com/TUMLegalTech/deconfounding_echr_emnlp22}{https://github.com/TUMLegalTech/deconfounding\_echr\_\\emnlp22}}
\end{itemize}

%% file: text/related.tex
LJP as an NLP task has been tackled using n-gram representations (e.g., \citealp{aletras2016predicting, medvedeva2020using}), word embeddings and domain models \cite{branting2021scalable}, and deep neural networks (e.g., \citealp{chalkidis-etal-2019-neural, ma2021legal, xu-etal-2020-distinguish}). Special attention must be given to the origin of the text from which the prediction is to be made. \citet{medvedeva2021automatic, medvedeva2022rethinking} recharacterize LJP on texts produced before the outcome is known as `forecasting' and observes that most current works `classify' judgments based on the data compiled after the outcome has been determined. They also find that forecasting is a harder task. This result is consistent with our finding of confounding effects from text production by the ECtHR, resulting in a prediction from fact descriptions that were influenced by the decision.

Moverover, the relationship between the information LJP models rely on and legal expert analysis of texts remains underexplored. \citet{bhambhoria2021investigating} find that transformer-based models exploit spurious correlations and that simple models, such as XGBoost, can achieve similar performance.
\citet{chalkidis2021paragraph} extract model rationales for alleged violation prediction and observes limited overlap with expert markup. Similarly, a small study in \citet{branting2021scalable} finds that users do not perceive case prediction-derived highlighting as useful in making predictions themselves. Our work contributes to this state of the art by using adversarial deconfounding to improve the overlap between what systems predict from with what legal experts consider relevant.

\noindent{\textbf{Deconfounding}} A growing number of works have raised awareness that deep neural models may exploit spurious statistical patterns and take erroneous shortcuts \cite{mccoy-etal-2019-right,bender-koller-2020-climbing,geirhos2020shortcut}. A common method of mitigating this is adversarial learning. \citealt{pryzant-etal-2018-deconfounded} use a gradient reversal layer \cite{ganin2016domain} to deconfound lexicons in text classification. Other domains that adopt adversarial training to eliminate confounders include bioinformatics \cite{dincer2020adversarial} and political science \cite{roberts2020adjusting}. Many existing works on identifying shortcuts focus on situations where these patterns are known in advance and may require potentially expensive data collection. In fairness-focused legal NLP, \citet{chalkidis-etal-2022-fairlex} observe and remedy group disparities in LJP performance on the ECtHR informed by metadata attributes (respondent state, applicant gender, applicant age). We extend this to explainability in LJP by involving a legal expert in a procedure that allows an efficient, incremental identification of distracting information, as well as its removal via adversarial training.

\noindent{\textbf{Interpretability}}
We employ interpretability techniques to evaluate model alignment with expert rationales. \citet{danilevsky-etal-2020-survey} reviews and categorizes the main current interpretability methods. Though initial works \cite{ghaeini2018interpreting, lee2017interactive} used attention scores as explanation for model decisions, \citet{bastings2020elephant, serrano2019attention} point out that saliency methods, such as gradient based methods \cite{sundararajan2017axiomatic, li2016visualizing}, propagation based methods \cite{bach2015pixel}, occlusion based methods \cite{zeiler2014visualizing}, and surrogate model based methods  \cite{ribeiro2016should} are better suited for explainability analysis. However, the reliability and informativeness of these methods remains an open research problem. Our model uses the currently most commonly used Integrated Gradients (IG) \cite{sundararajan2017axiomatic}, which computes the gradient of the model’s output with respect to its input features.

%% file: text/dataset.tex
The ECtHR has been the subject of substantial prior work in LJP. We use two datasets for model training and evaluation: First, for \textbf{binary violation} we use the dataset by \citet{chalkidis-etal-2019-neural} of approx. 11k case fact statements, where the target is to predict whether the court has found at least one convention article to be violated. To evaluate alignment, we annotate 50 (25 each) expert rationales for cases from both the development and test partitions (See App. \ref{app:annotationProcess} for the annotation process). Second, for \textbf{article-specific violation}, we use the LexGLUE dataset by \citet{chalkidis-etal-2022-lexglue}, which consists of 11k case fact statements along with information about which convention articles have been alleged to be violated, and which the court has found to be violated. For alignment, we merge this data with the 50 test set rationales from \citet{chalkidis2021paragraph}. While both datasets stem from the ECtHR’s public database, they differ in case facts and outcome distribution as we explain in Sec. \ref{sec:data-distribution}. The input texts consist of each case's FACTS section extracted from ECtHR judgments. This section is drafted by court staff over the course of the case proceedings. While it does not contain the outcome explicitly, it is not finalized before the final decision has been determined, potentially creating confounding effects.

We conduct experiments on four LJP tasks:

\noindent{\textbf{Task J - Binary Violation}}
For our task \textbf{J}, the model is given a fact statement and is asked to predict whether or not any article of the convention has been violated. We train our models on \citet{chalkidis-etal-2019-neural} and evaluate alignment on the set of expert rationales we collected.

\noindent{\textbf{Task B - Article Allegation}}
We train and evaluate on LexGLUE's \textit{ECtHR B},\footnote{The LexGLUE dataset does not contain metadata (case id, Respondent state etc); in this work we use an \href{https://www.kaggle.com/datasets/mathurinache/ecthrnaacl2021}{enriched version} of the same dataset by Mathurin Aché.}
where the fact description is the basis to predict the set of convention articles that the claimant alleges to have been violated. It can be conceptualized as topic classification in that the system needs to identify suitable candidate articles (e.g., the right to respect for private and family life) from fact statements (e.g., about government surveillance). We test alignment on the expert rationales by \citet{chalkidis2021paragraph}.

\noindent{\textbf{Task A - Article Violation}}
We also experiment with LexGLUE's \textit{ECtHR A}, which is to predict which of the convention's articles has been deemed violated by the court from a case's fact description. Task A is a more difficult version of task B, where both an identification of suitable articles and a prediction of their violation must be performed. For alignment, we again use the expert rationales by \citet{chalkidis2021paragraph}, which are technically intended for task \textit{ECtHR B}, but which we consider to also be suitable for an evaluation of task A.\footnote{The annotation explanations in \cite{chalkidis2021paragraph} state that ``\textit{The annotator selects the factual paragraphs
that ``clearly'' indicate allegations for the selected article(s)}''. We hypothesize that the so annotated passages contain information that is legally relevant for the violation as well.}

\noindent{\textbf{Task A|B - Article Violation given Allegation}}
We further disentangle the LexGLUE tasks and pose \textit{ECtHR A|B}. Given the facts of a case and the allegedly violated articles, the model should predict which (if any) specific articles have been violated. This task reflects the legal process, as the court is aware of allegations made by the applicants when deciding. Providing information about the allegations shifts the nature of the task from topic classification to article-specific violation/non-violation prediction, thus refocusing the model and ideally leading to violation-specific explanations.



\subsection{Data Distribution \& Preprocessing}
\label{sec:data-distribution}
In order to facilitate model alignment, we worked with our ECtHR expert to identify shallow prediction signals in the fact statements that are unrelated to the legal merits of the complaint.

\subsubsection{Length and Respondent State}
For the task J dataset of \citealt{chalkidis-etal-2019-neural}, we find that the distribution of fact description length (number of sentences) and the distribution of respondent states are different between the two classes (see Appendix  \ref{app:dataset-stats}). We hence account for the identity of the respondent state and the length of the fact descriptions via our deconfounding procedure for both datasets.

\subsubsection{Accounting for Inadmissible Cases}
We also observe in the task J dataset that the magnitudes of the running paragraph numbers differ between the classes, and that the single word ``represented'' strongly correlates with the positive class. This phenomenon arises because 2.6k of the 7k training cases are `inadmissible' cases labeled as ‘non-violation’. Legally, inadmissible cases are not necessarily ‘non-violation’ as inadmissibility relates to complaints not fulfilling the court's formal or procedural criteria.\footnote{For example, the applicants lodge the complaint outside the time limit after the final domestic judicial decision or fail to exhaust required domestic remedies before complaining to the ECtHR, etc. It should be noted that the majority of inadmissible cases are decided by single judges and not available on the public database \href{https://hudoc.echr.coe.int}{HUDOC}.} In such cases, the court does not examine the merits of the application. The more interesting non-violation cases are such that are admissible, but in which no violation of the convention has been found. The single negative class contains instances of both inadmissible and admissible-but-no-violation-found cases. As explained above, the input texts of \citealt{chalkidis-etal-2019-neural} are extracted from the FACTS section of full ECtHR decisions. In inadmissible cases, the applicant's background information can typically be found at the beginning of that section. We found that almost all inadmissible case facts start with the same formulaic sentence stating the applicant's name, nationality, and legal representation. This specific sentence is absent from the texts of admissible cases (violation and non-violation), where that information is part of a separate PROCEDURE section not included in the dataset. Moreover, due to the PROCEDURE section preceding the FACTS section in admissible cases, the running paragraph numbers appearing in FACTS sections of inadmissible cases are smaller than those of the admissible cases. If not remedied, these phenomena provide a considerable predictive signal for the label and distract the system from legally relevant information. In our experiments, we hence remove paragraph numbers from the input via preprocessing and account for distractor vocabulary via our deconfounding procedure described in Sec. \ref{sec:deconfounding}. Still, the nature of task J remains unchanged and requires the system to classify the outcomes of a collection of both admissible and inadmissible cases.

\subsubsection{Article-Specific Violation}
\label{sec:article-specific-violation}
By contrast, the more recent LexGLUE dataset only contains admissible cases and corresponding information about which articles the claimant has alleged to have been violated (for task B) along with those that the court has found to have been violated, if any (task A). The collection covers 10 different convention articles that make up the largest share of ECtHR jurisprudence. Each article has been alleged in a partition of the cases, and has been found to be violated in a subset of these.\footnote{A few cases exist where the court refocuses the issues and finds a violation of an article that has not been alleged, but in the dataset they only occur in a negligibly small number of instances. (see App. Sec. \ref{app:lexglue-char})} For a given article in task B, all cases in which it has been alleged can be considered positive instances while the remaining cases are negatives. We consider task B as akin to topic classification, where the rights enshrined in the convention articles (e.g., Art. 6: right to a fair trial; Art. 1 Protocol 1: protection of property, etc.) may correlate with certain case fact language (e.g., related to law enforcement or expropriation, respectively). Task A incorporates this step and adds violation prediction per article, which is more difficult in principle. However, we observe that a few articles account for a large portion of the data and the conditional probability of a positive violation label in task A given its allegation labels from task B can be very high (see App. \ref{app:lexglue-char}). This makes an analysis of what trained models focus on more difficult, since they may learn to identify these dominant articles with high conditional violation probability, and be distracted from focusing on information that specifically signals violations of those articles. To remedy this, we propose task A|B that provides models an easy access to the label information of B, facilitating their focus only on determining whether the court finds a violation of given articles. This task is realistic since the allegations by the claimant are known to the court at the time that it decides whether the respondent state has violated the convention in the case.



%% file: text/framework.tex
We apply an expert-informed deconfounding method designed to mitigate the distracting effects of confounding elements and spurious correlations. As \citet{pryzant-etal-2018-deconfounded} observe, accounting for confounders is common practice throughout many data analysis tasks to capture the intended signal and facilitate explainable models. In LJP, we understand confounding elements as such that influence both the observed legal outcome (convention violations found by the court as coded in the dataset) and the input text from which this outcome is to be predicted (here: ECtHR fact statements). Already covered examples are the different distribution of information across sections for admissible and inadmissible cases and the length of the fact descriptions (inadmissible cases tend to require less factual information to be dismissed).\footnote{If one assumes sectioning to be dependent on the case outcome, then this could even be characterized as an inverse causality relationship. For simplicity, and to account for court-internal document production processes, we understand ``confounder'' as including such configurations.} An example of a spurious correlation is the identity of the respondent state (certain article violations will be claimed more often against a small number of governments, leading to a correlation). They each should have no bearing on the probability of an outcome in a given case as a judge will not decide against a violation because the facts are short, or because the case is against a particular government.

Confounding effects and spurious information in LJP may not be known ahead of time, especially if the legal decision is not made on the basis of an immutable a priori document, but rather on the basis of text that is technically a part of the eventual judgment. 
Our expert-informed method is intended to mitigate such situations where spurious correlations are introduced in the text production but may not be known in advance as explicit confounders.

Our method consists of two steps: (i) Identification of distracting attributes for deconfounding through a combination of simple model training and minimal expert markup, and (ii) mitigation of these effects through adversarial training.

\subsection{Step 1: Identification of Distracting Attributes and Tokens}
We first identify input attributes and categorize them as either distracting or genuinely legally relevant in an expert consultation. `Distracting' attributes are highly correlated with the task label but not relevant in a human expert prediction. Attributes can be either  (i) explicit in the text (such as vocabulary tokens) or (ii) implicit (e.g., country, text length, etc.). Implicit attributes can be derived from available metadata or a corpus analysis.

For textual attributes, we apply depth-limited decision trees on an n-gram representation of the fact statement to predict the case outcome. We extract all tokens that appear in the trees and iterate, successively removing tokens identified as predictive. Compared to extracting tokens from a single larger tree, this process is better suited to remove high-entropy-reducing tokens one typically finds near the root of trees. The list of removed tokens is then presented to a legal expert, who categorizes them into spurious and legally genuine (see Appendix Sec. \ref{app:spurious-vocab} for the list of spurious vocabulary identified by the expert and the rationale behind the choices). This requires substantially less effort from the expert compared to other methods, such as data annotation or manual creation of counterfactuals. To prevent trees from picking up very sparse tokens, we filter the extracted terms using local mutual information (LMI) \cite{schuster2019towards}, a re-weighted version of pointwise mutual information (PMI) \cite{church1990word}. We calculate LMI for each pair of token and label as illustrated in Appendix Sec. \ref{app:lmi}. 

\subsection{Step 2: Mitigation of Distracting Attributes}
We assume a neural NLP model $M$ consisting of a feature extractor $F$ and classifier $C$ with parameters $\theta_f$ and $\theta_c$, respectively. For each confounder $k$, we apply a discriminator $D_k$ with parameters $\theta_{d_K}$ to the feature extractors. We use adversarial training to maximize the feature extractor's ability to capture information for the main classification target while minimizing its ability to predict the value of distractor attributes. This encourages the model to generate distractor-invariant feature representation for the classifier. We use the following adversarial training objective: 

\begin{equation}
\scalemath{0.75}{
  \sum\limits_{k} \arg \min\limits_{\theta_{d_k}} L(D_k(F(x)), y_k)}
\end{equation}
\begin{equation}
\scalemath{0.75}{
    \begin{aligned}
   \arg \min\limits_{\theta_{f}, \theta_{c}} ~~ &   [L(C(F(x)), y_c) -  \sum\limits_{k} \lambda_k L(D_k(F(X)), y_k)]
    \end{aligned}}
\end{equation}
where $L$ represents the loss, $\lambda$ is a hyperparameter, $x$ is the input, $y_c$ is the label, and $y_k$ is the distracting attribute $k$. The above optimization is performed using a gradient reversal layer (GRL) \cite{ganin2015unsupervised} to jointly optimize all the components instead of alternately updating the components as in GANs \cite{goodfellow2014generative}. The GRL is inserted between the feature extractor and discriminators. It acts as the identity during the forward pass but, during the backward pass, scales the gradients flowing through by $-\lambda$, making the feature extractor receive the opposite gradients from the discriminator. This changes the overall objective function to :
\begin{equation}
\scalemath{0.75}{
    \begin{aligned}
   \arg \min\limits_{\theta_{f}, \theta_{c}, \theta_{D}}    [L(C(F(x)), y_c) +  \sum\limits_{k} \lambda_k L(D_k(GRL((F(X))), y_k)]
    \end{aligned}}
\end{equation}

We hypothesize that learning distractor-invariant feature representations through adversarial learning will help the model to focus on parts of the input that experts consider relevant.

%% file: text/experiments.tex
In this section we describe our experiments in using our proposed deconfounding methodology to improve the alignment of model focus on the input with expert rationales on our set of LJP tasks.

\subsection{Models}
{\bf Baseline}: We use the BERT variant of Hierarchical Attention Networks \cite{yang2016hierarchical} as a baseline model. To segment our very long input texts we resort to a greedy sentence packing strategy in which we pack as many sentences as possible into one packet until it reaches the predefined maximum length (512 tokens constrained BERT). When a sentence exceeds this maximum, we split it into parts to fit into multiple packets.
We encode each packet with LegalBERT \cite{chalkidis2020legal} to obtain the token level representations. Following \citealp{yang2016hierarchical}, we use a token attention layer aggregating the representation of the tokens and form a sentence (packet) vector. We pass these sentence vectors through a GRU encoder to obtain contextual representations. These are aggregated at the document level using a sentence attention layer. This model constitutes the feature extractor component $F$ in our architecture. The obtained document representation is passed through dense layers for the final target prediction, constituting our classifier component $C$. 

\noindent {\bf paraRem}: Same as the baseline model but trained on data from which the paragraph number artifacts have been removed (see Sec. \ref{sec:data-distribution}).

\noindent {\bf gradCou}: \textit{paraRem} model extended with a multi-class discriminator with a cross-entropy loss predicting the identity of the respondent government, and a corresponding deconfounding GRL.

\noindent {\bf gradLen}: \textit{paraRem} model extended with a length discriminator predicting the length (number of sentences) of the document via a set of bins and a cross-entropy loss, and a corresponding GRL to predict the bin value. 

\noindent{\bf gradVocab}: \textit{paraRem} model extended with a vocabulary discriminator to predict the presence of identified spurious tokens, and associated GRL. As there can be multiple spurious tokens in a document, we employ binary entropy loss per token as it is a multi-label classification.

\noindent We refer to the above three deconfounded models collectively as \textit{singleGrad} models.


\noindent {\bf gradAll}: \textit{paraRem} model extended with all country, length, and vocabulary discriminators in parallel, and associated GRLs.

\noindent Please refer to Appendix Sec. \ref{app:model-config} for details on model configuration and training. 

\subsection{Quantitative Evaluation \& Discussion}
\subsubsection{Expert Alignment Evaluation}

Our main objective is to evaluate the alignment of the model's focus on the input text with legal expert rationales (i.e., selected subsets of relevant segments of the input). Following \citealt{chalkidis2021paragraph}, we measure the model's ability to identify the correct rationales at the paragraph level, which is the natural granularity of ECtHR fact sections. To extract the importance score for each paragraph, we rely on an interpretability technique which quantifies the impact of a particular input token towards the final prediction of the model.

We use integrated gradients \cite{sundararajan2017axiomatic} to obtain a token-level focus score and aggregate paragraph-level scores as the squared L2-norm of token scores in the paragraph divided by the square root of its number of tokens to account for length variation. We compute precision@k conditioned on some fixed $k$ between the top-k paragraphs based on paragraph scores and golden paragraph rationales. The number of relevant paragraphs in gold rationales varies considerably, so a predefined $k$ is inadequate. Thus, we compute precision@Oracle following \citealp{chalkidis2021paragraph}, where \textit{Oracle} is the number of relevant paragraphs in the gold rationales.

For tasks J, A, and A|B, the negative label (i.e., non-violation) is of similar interest as the positive label. In task B, however, the  negative label merely indicates that a specific article has not been alleged, which is legally largely uninteresting. Hence, we reduce negative IG scores of tokens (indicating a negative contribution to the prediction) to zero. 

\begin{table}
      \centering
        \scalebox{0.9}{
         \begin{tabular}{|c|c|c|c|c|c|}
  \hline
    &  \multicolumn{2}{c|}{\textbf{J}} & \multicolumn{1}{c|}{\textbf{B}} & \multicolumn{1}{c|}{\textbf{A}} & \multicolumn{1}{c|}{\textbf{A|B}} \\
  \hline
   \textbf{Model} & \textbf{valid} &\textbf{test} & \textbf{test} &\textbf{test} &  \textbf{test} \\ 
 \hline
 \multirow{2}{*}{Random}  & 38.67 & 28.22 & 36.65 & 36.65 & 36.65 \\
 & \footnotesize(4.52) & \footnotesize(3.16) & \footnotesize(2.91) &  \footnotesize(2.91) &  \footnotesize(2.91) \\  \hline
 \multirow{2}{*}{baseline} & 39.04 & 29.73 & 39.07 & 40.10 & 41.36 \\
 & \footnotesize(4.31) & \footnotesize(3.28) & \footnotesize(2.94) & \footnotesize(3.02) & \footnotesize(3.02) \\ \hline
 \multirow{2}{*}{paraRem} & 41.81 & 31.53 & 41.47 & 41.93 & 41.86 \\
 & \footnotesize(3.59) & \footnotesize(3.43) & \footnotesize(3.06) & \footnotesize(2.88) & \footnotesize(2.86) \\ \hline

 \multirow{2}{*}{gradCou} & 42.29 & 33.37 & 42.36 & 43.56 & 43.16 \\
 & \footnotesize(3.54) & \footnotesize(3.75) & \footnotesize(3.09) & \footnotesize(2.78) & \footnotesize(2.85) \\ 
 \multirow{2}{*}{gradLen} & 42.21 & 33.58 & \emph{43.55} & 43.12 & 43.75 \\
 & \footnotesize(3.55) & \footnotesize(2.98) & \footnotesize(2.92) & \footnotesize(3.31) & \footnotesize(3.14) \\
 \multirow{2}{*}{gradVocab} & \emph{42.96} & \emph{33.77} & 43.34 & \emph{44.48} & \emph{44.39} \\
 & \footnotesize(3.57) & \footnotesize(3.41) & \footnotesize(3.02) & \footnotesize(3.36) & \footnotesize(2.86) \\ \hline
 \multirow{2}{*}{gradAll} & \textbf{44.42} & \textbf{34.48} & \textbf{44.84} & \textbf{45.95} & \textbf{44.91} \\
 & \footnotesize(3.40) & \footnotesize(3.74) & \footnotesize(3.13) & \footnotesize(3.09) & \footnotesize(3.18)  \\ 
 p-value & \footnotesize 0.164 &  \footnotesize 0.013 & \footnotesize 0.031 &  \footnotesize 0.008 &  \footnotesize 0.071 \\ \hline

  \end{tabular}}
        \caption{Expert alignment performance expressed in precision@Oracle scores; value in brackets indicates standard error of the computed score; p values compare gradAll versus paraRem and were computed using a paired t-test.}
  \label{tab1}
\end{table}

\subsubsection{Prediction Performance Evaluation}
We also report the models' performance on the main four LJP tasks. For Task J, we report the macro F1-score for binary violation prediction. For Task A and B, following \cite{chalkidis-etal-2022-lexglue}, we report micro-F1 ($\mu$-F1) and macro-F1 (m-F1) scores. For Task A|B, we also report micro-F1 and macro-F1 scores. In computing the above metrics for tasks A and A|B, we consider the cases in which a particular article has been deemed violated as positive instances and the rest of the instances as negatives. We also introduce \textit{hard-macro-F1} (hm-F1) for both Task A and A|B, in which F1 is computed for each article using only those instances as negatives where an article has been alleged as violated but not found so by the court. 

\subsubsection{Quantitative Evaluation Results}
Table \ref{tab1} and Table \ref{tab2} show the performance of  different models on expert alignment and outcome prediction, respectively. 

\noindent {\bf paraRem vs. baseline}: We observe that paraRem outperforms the baseline model in expert alignment across all tasks with a minimal drop in prediction performance. Task J stands out in that removing distracting signals via paragraph number removal even leads to a marginal improvement. Notably, we separately confirm the vulnerability of the baseline model by applying it to the test set with paragraph numbers removed and evaluate it on a test set without paragraph numbers, resulting in macro-F1 of 51.16 (i.e., a nearly 30 points drop). 

\noindent {\bf gradCou, gradLen, gradVocab vs. paraRem}: In all tasks, we observe that all singleGrad models improve in expert alignment performance over paraRem by a small but consistent margin. This demonstrates the ability of our deconfounding component to help the model better identify legally relevant parts of the input. Notably, gradVocab shows the most improvement in alignment over paraRem in all tasks except Task B (alleged article prediction), where gradLen performs best. During development on task B, we observed that the decision-tree based removal of predictive words led to only a marginal falloff in tree model accuracy, even after multiple iterations, since there was simply a lot of topical words (e.g., for police misconduct, legal proceedings, etc.) to take over as some of them were removed. This in part reflects the different nature of the tasks and shows a limitation of our tree-training-based method for identifying spurious tokens. Similar to paraRem, the gradLen model (in case of Task A, B, and A|B) also shows improvement in prediction performance compared to the baseline model. This suggests that deconfounding can potentially  prevent the model getting stuck in distractor-related local optima.

\noindent {\bf Alignment}: All singleGrad models outperform the baseline with regard to expert alignment. We observe that gradAll achieves the highest score, which establishes some degree of complementarity among the three singleGrad models and the distracting signals they remedy. A paired t-test (gradAll vs. paraRem) reveals p-values above typical significance levels for the validation partition of task J, along with a considerable divergence in the general score level for the two tasks. We conjecture that this is the result of our small rationale sample size (50 from each partition) and differences in distribution between the task J data partitions, which have been split along the timeline rather than random. We also see a higher p-value for task A|B, which is intuitive since it is the most difficult. Its dataset lacks easily identifiable inadmissible cases (as in task J) and it has access to B's labels as concurrent, non-textual input. To gain some more insight into A|B, we report on a qualitative error analysis of the model rationales below.

\begin{table*}
      \centering
            \scalebox{0.9}{ \begin{tabular}{|c|c|c|c|c|c|c|c|c|c|}
  \hline
   &  \multicolumn{1}{c|}{\textbf{Task J}} & \multicolumn{2}{c|}{\textbf{Task B}} & \multicolumn{3}{c|}{\textbf{Task A}} & \multicolumn{3}{c|}{\textbf{Task A|B}} \\
  \hline
   \textbf{Model} & \textbf{m-F1} &\textbf{$\mu$ F1} & \textbf{m-F1} &\textbf{$\mu$ F1} &  \textbf{m-F1} &  \textbf{hm-F1} & \textbf{$\mu$ F1} &  \textbf{m-F1} &  \textbf{hm-F1} \\ 
 \hline
 baseline & 81.23 &  \textbf{78.08} & \textbf{68.42} & \textbf{69.28} & \emph{58.80} & 55.30 & 77.42 & 68.23 &  58.95  \\ \hline
 paraRem & \textbf{82.67} & 77.82 & 66.90 & 68.94 & 58.35 &  55.62 & 77.20 & 67.32 & 58.80 \\ \hline
 gradCou & 81.22 & 76.31 & 66.58 & 67.47 & 56.40 &  54.04 & 76.83 & 66.85 & 58.14 \\
 gradLen & 81.99 & \emph{78.06} & \emph{67.19} & \emph{69.18} & \textbf{58.88} & \textbf{56.10} & \textbf{79.07} & \textbf{69.79} &  \textbf{61.24} \\
 gradVocab & \emph{82.00} & 77.68 & 66.40 & 69.06 & 58.47 & \emph{55.71} & \emph{78.87} & \emph{69.49} & \emph{60.64} \\ \hline
 gradAll & 81.53 & 77.46 & 66.75 & 68.32 & 58.06 & 53.74 & 78.71 & 69.26 & 60.58 \\
 \hline
  \end{tabular}}
        \caption{Prediction Performance}
  \label{tab2}
\end{table*}

\subsection{Qualitative Evaluation \& Discussion}
\noindent {\bf Expert Scores}: We sample 40 cases from task A|B validation and test sets (see App. Sec. \ref{app:case-sampleing}). We provide the expert with randomized visualizations of IG scores at the token level derived from our paraRem and gradAll models. Following \cite{jayaram2021human}, the expert was asked to rate these on a five-point Likert scale (range -2 to 2) on two metrics: (i) Sufficiency: Is a sufficiently large set of tokens focused on to arrive at the prediction?; and (ii) Irrelevance: How many irrelevant tokens does the model focus on? We phrased the scale such that, for both parameters, a higher rating signals a better alignment between the model focus and the expert's assessment. Table \ref{tab3} presents averages of the raw scale scores. We observe that the deconfounded gradAll model scores higher (See App. \ref{app:screenshots} for an example pair of IG visualizations).

\noindent {\bf Manual IG Inspection}: For the paraRem model, we notice that high scoring IG tokens are sparse, whereas in gradAll, focus is densely distributed. There, contiguous spans of tokens tend to receive higher scores. This phenomenon is likely due to paraRem being drawn to single word distractors. Deconfounding helps the gradAll model to spread its focus across larger segments of the text. Our ECtHR expert further observed that gradAll highlighted words that, in conjunction, were indicative of the outcome, even if those were a considerable distance apart. At the same time, however, it seemed that two words hinting at opposite outcomes in a single sentence forced the system to focus only on one of the two, leaving the other one unhighlighted. We conjecture that these long- and short-distance phenomena are a result of the hierarchical model architecture necessitated by the long documents and leave their further exploration for future work.

An inspection of high scored tokens in paraRem reveals that many of them are highly discriminative in our decision tree models, showing that complex neural models can easily fall for distractors at the expense of missing equally predictive but semantically more complex signals. This reinforces our paradigm to identify discriminative tokens using a simpler model and subject them to expert scrutiny. In particular, we found that the word ``represented'' forms a natural decoy and, when injected into a violation-outcome fact statement, flips the predicted label of trained deep neural models. This led us to believe those models rely more on individual words than one might expect, and motivated us to explore how this can be exploited with information derived from simple models.
Figure \ref{fig:dt} shows that the performance of decision trees with unigram features (at iteration 1 without removed tokens) can even come close to BERT models. 

In paraRem, we further observe that tokens at the start of sentences receive higher IG scores. We believe this to be the model counting sentences, which justifies deconfounding for length. For gradAll, we observe that sentence beginnings still receive focus, but less strongly so. This may be due to BERT recognizing sentence boundaries.


\begin{table}[]
    \centering
        \scalebox{0.9}{
    \begin{tabular}{|c|c|c|}
      \hline
   \textbf{Metric} & \textbf{paraRem} &\textbf{gradAll} \\  
   \hline
    Sufficiency ($\uparrow$) & 0.150 & \textbf{0.550} \\
    Irrelevance ($\uparrow$) & 0.475 & \textbf{0.625}    \\
    \hline
    \end{tabular}}
    \caption{Qualitative evaluation scores}
    \label{tab3}
\end{table}

\noindent {\bf Further alignment improvement}: The overall low precision@Oracle scores show that considerable differences in alignment with human experts remain. We conjecture that the model is shifting its focus, at least in part, to other spurious attributes which our current setup could not reveal. This calls for further investigation to design effective methods to identify such patterns. However, we expect them to be increasingly subtle and difficult to recognize, potentially even for legal experts. An intuitive upper bound for the system would be the annotation agreement of multiple experts, which to the best of our knowledge remains unexplored in the current state of the art.

\noindent {\bf Expert Pattern Identification}: Our results naturally raise the question of how distractors can be identified in ECtHR fact texts by experts. Generally, the patterns we focused on affect the relationship between the argumentation in the judgment and the supportive facts given. There is copious literature on the court’s inconsistent approach to legal decision-making (e.g., \citealt{madsen2018backlash}) and it is known to switch between judicial policies depending on case circumstances \cite{helfer2020walking}. We hence paid attention to specific markers in the fact section and correlated them to existing precedents and argumentation patterns. A few examples: The court may decide to make use of positive obligations and decide against the state (violation) by highlighting failures of national authorities, or may decide to use those same positive obligations under ‘the responsible authorities’ doctrine, highlighting the efforts of national authorities to bring domestic legislation in line with the convention, thus deciding that there has been no violation. There are also fact patterns and practices specific to particular state parties to the convention (e.g., prison overcrowding, procedural issues in child abduction cases). The court may also sometimes highlight specific facts of a case with the view to ‘document’ its resemblance to, or divergence from, an existing precedent. A detailed, legally informed case study on predictive patterns is beyond the scope of this work.

\subsection{Recommendations for LJP Research}
\label{ljp_recommendations}
In order to produce value for legal practice, we believe that LJP/LJF as an NLP task should strive for a productive combination of expert knowledge with data-derived insight. Based on our results, we formulate the following recommendations: First, as has already been observed in the field, any prediction/classification should happen from suitable source text that does not encode information about the outcome but contains as complete factual information as possible, or at least control for this influence. Second, straightforward predictors (e.g., input length and shallow unigram models) should be used to identify distractors and confounders. Third, claimed performance levels in predicting case outcomes should be contextualized by information about the distribution of the legal issues and respective conditional outcome probabilities in the corpus, as well as against baseline classifiers capable of exploiting known distractors. Fourth, more granular outcome variable information (e.g., case declared inadmissible vs. case dismissed on the merits, decomposition into outcomes of individual issues) will allow the development of more nuanced prediction/classification systems. Taken together, if such models can be explained and integrated into a decision support system for suitable tasks in legal practice, experts will be more likely to perceive them as adding value.

%% file: text/conclusion.tex
Our results show that our deconfounded LJP models are consistently better aligned with expert rationales than a baseline optimized for the target label only, and in many cases can even achieve better prediction performance. However, the improvement is small and the paragraphs focused on by all our models are still quite different from what an expert has annotated as relevant, as indicated by generally low precision@Oracle scores (<50\%). Still, our quantitative results show that expert-informed deconfounding LJP works in principle and can potentially go a long way to train more robust and trustworthy neural LJP models, as well as derive more useful legal insight from them.

%% file: text/limitations.tex



We present a case study in deconfounding legal judgment prediction on the ECtHR, and all results are to be understood as relative to the ECtHR, its jurisprudence, the used datasets, and the formal tasks. The distracting attributes we identify include confounding effects of the court's document production, where the decision may be known before the decision text (including the fact section) is finalized. A replication of this study in other LJP settings is of course warranted before general applicability can be claimed. Our analysis of task B has further revealed that redundant vocabulary distribution can challenge the system's ability to point out individual `smoking gun' distracting tokens. This aspect is particularly complex in light of differing legal systems and their respective cultures and patterns of drafting texts that may form the basis of predictive or, more generally, assistive systems. Morphologically rich languages, where distracting signal may be spread across multiple tokens, may make this challenge more difficult and require stem- or lemma-based processing as part of the method.

Our deconfounding method is work-intensive and assumes the identifiability of distracting information in text and metadata by an expert. Legal expert agreement about what parts of decisions are relevant remains underexplored, and the division of genuine versus spurious language may also vary in between multiple experts. While we are convinced that further research on effective deconfounding of legal NLP systems is needed if these systems are to become robust and trustworthy, the time-intensive nature of collaboratively developing and qualitatively evaluating such models with legal experts poses a considerable resource challenge.

A technical difficulty in working with legal documents is their length, and the use of packet-based hierarchical models constrains the maximum distance across which tokens can directly attend to one another. The impact of this limitation on model performance in various types of tasks is the subject of ongoing exploratory work (e.g., \citealt{dai2022revisiting}).

%% file: text/appendix.tex
\section{Dataset Statistics}
    
\label{app:dataset-stats}
\begin{figure*}[]
\centering
\includegraphics[width =\textwidth]{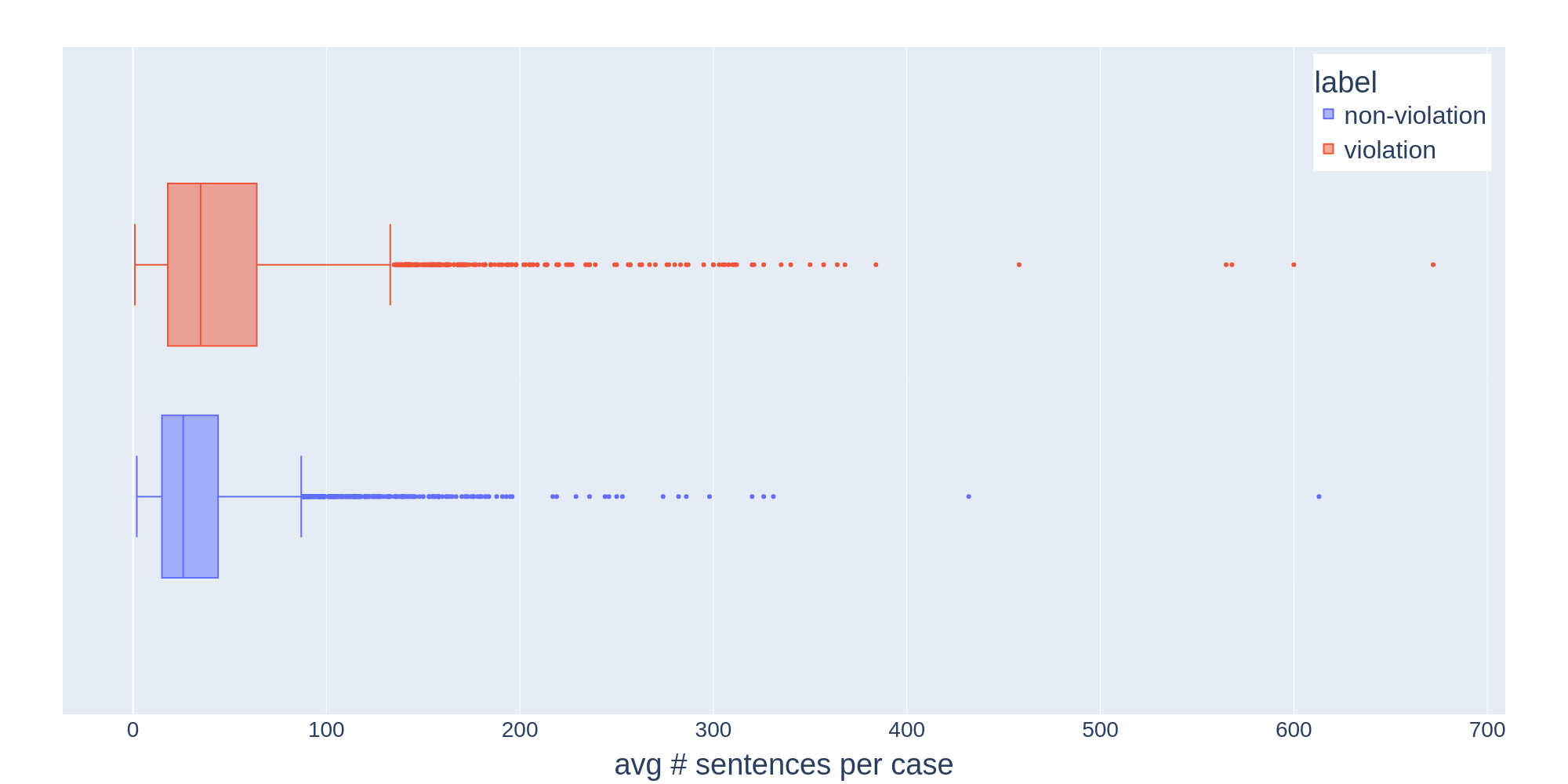}
{\caption{\label{len_train_j} Text length distribution in the training set of Task J}}
\end{figure*}

\begin{figure*}[]
\centering
\includegraphics[width =\textwidth]{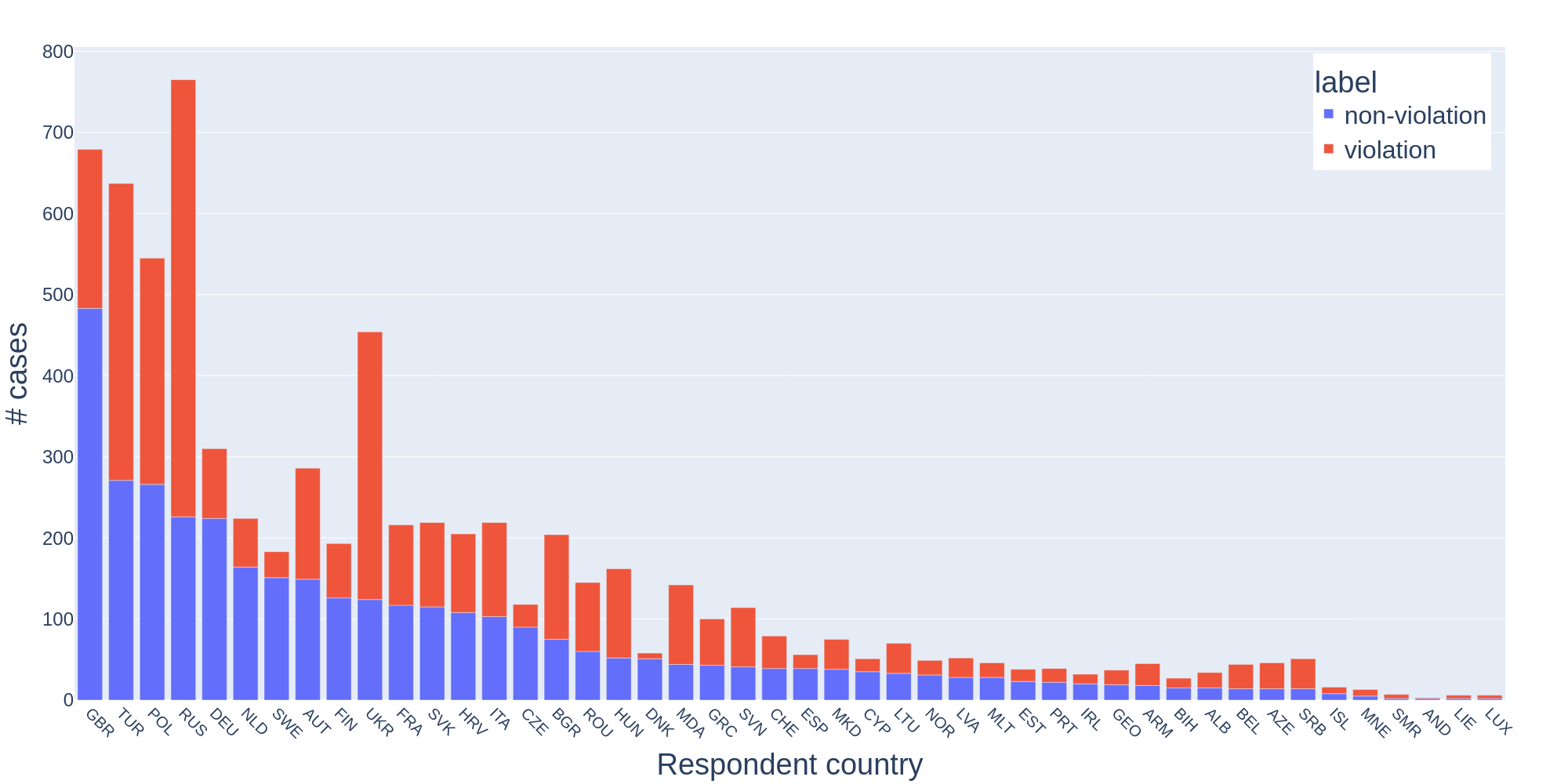}
{\caption{\label{cou_train_j} Country distribution in the training set of Task J}}
\end{figure*}

\begin{figure*}[]
\centering
\includegraphics[width =\textwidth]{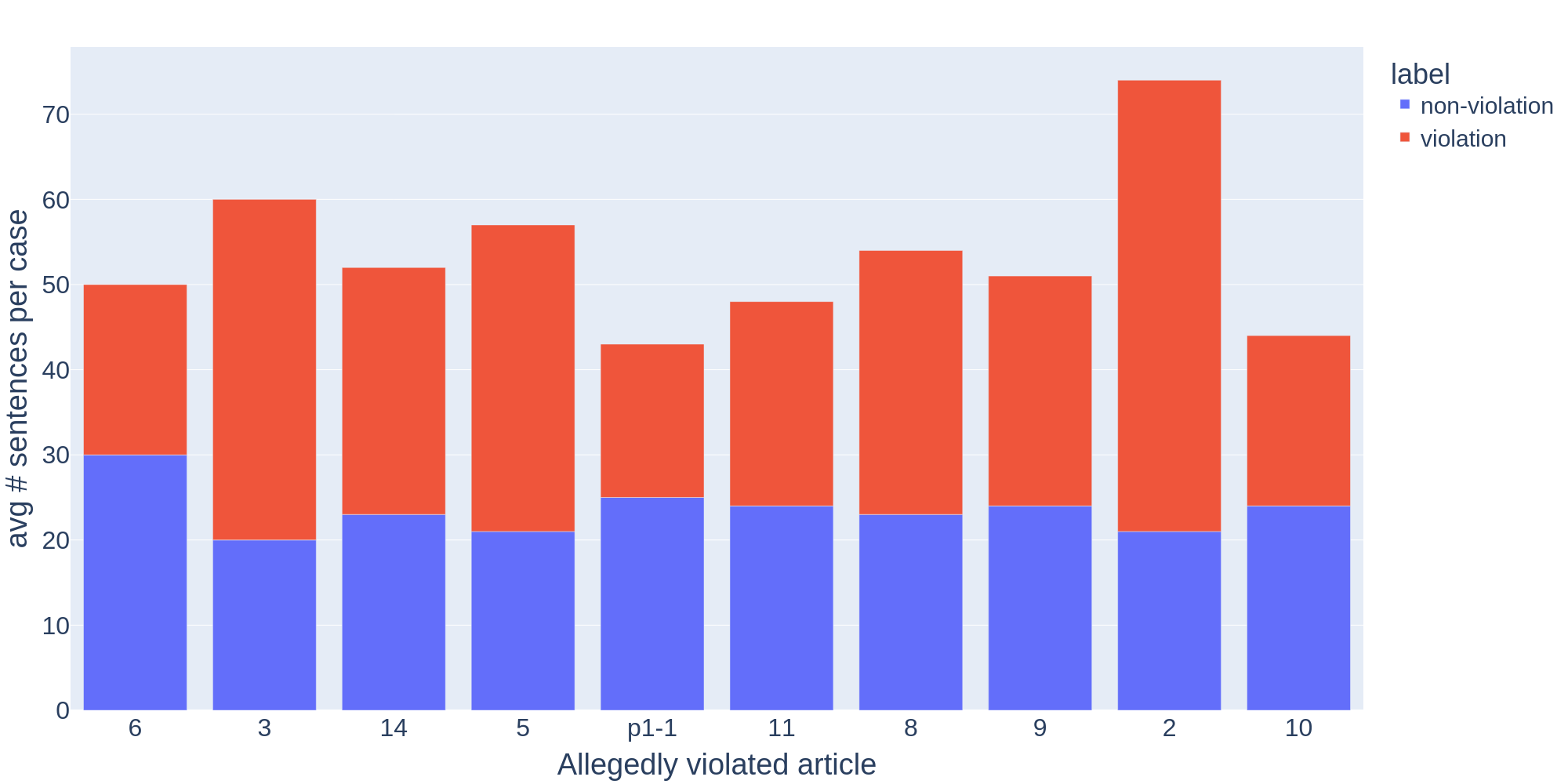}
{\caption{\label{len_b_train} Text length distribution for each violated article in the training set of Task B}}
\end{figure*}

\begin{figure*}[]
\centering
\includegraphics[width =\textwidth]{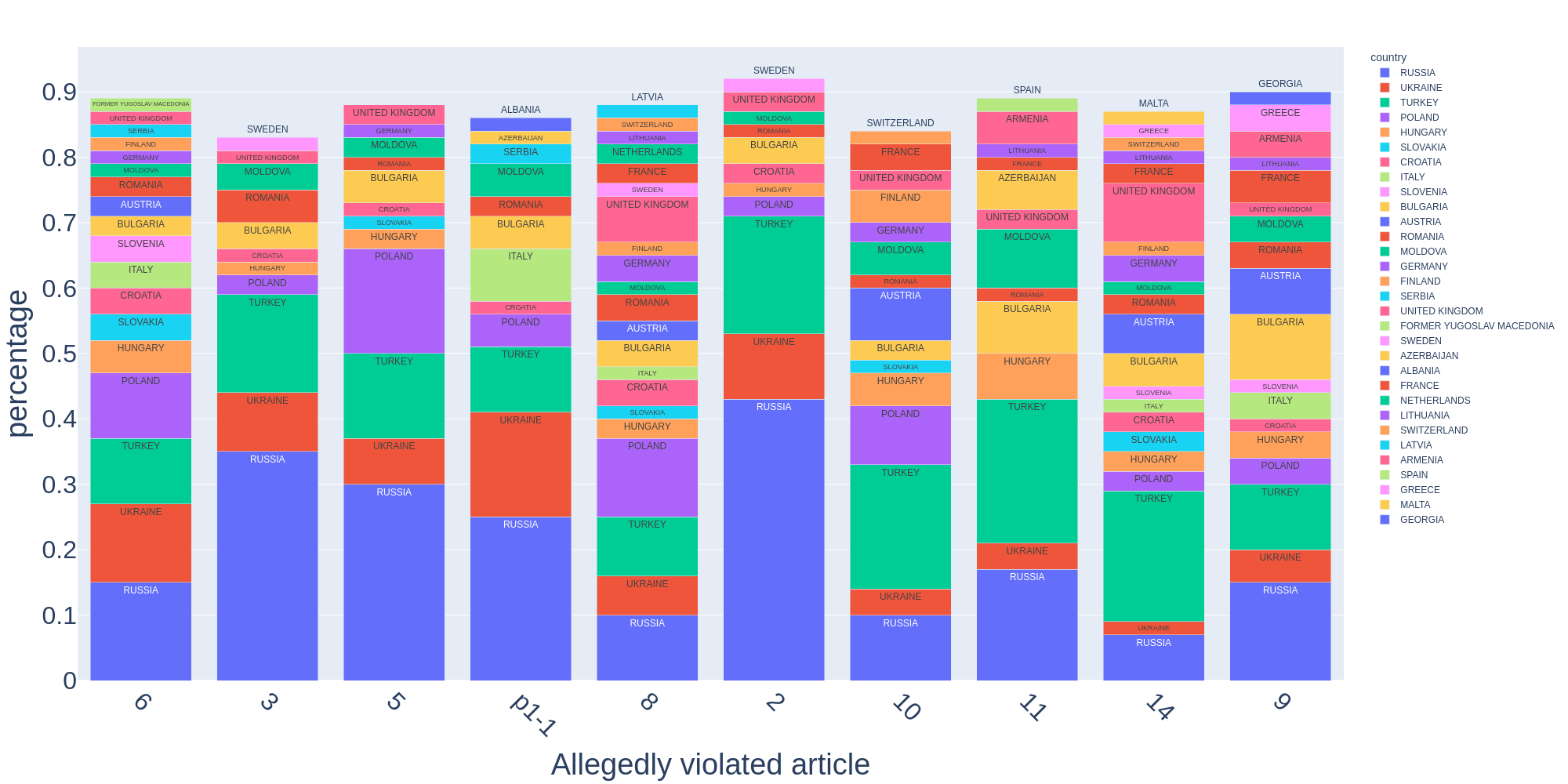}
{\caption{\label{cou_b_train} Country distribution in training set of Task B. For the display effect, only countries accounting for more than 1\% of the total cases are displayed. }}
\end{figure*}

Table \ref{artefactsJ} demonstrates the artefacts from corpus construction and admissibility-related confounding information in the training set of Task J. Figure \ref{len_train_j} and \ref{cou_train_j} display the distribution of text length and the respondent state in the Task J train set. Figure \ref{len_b_train} and \ref{cou_b_train} show the statistics of text length and respondent state in the Task B train set.

\section{LexGlue Dataset Characteristics}
\label{app:lexglue-char}   
Table \ref{lexglue-stats} describes the conditional probability of a violation finding by the court given the allegation of a particular article as well as the probability of a violation finding regarding a particular article even though it was not alleged. 

\section{Rational annotation Process for Task J}
\label{app:annotationProcess}
We sampled 50 cases (25 each) from the validation and test split. In each split, we sample two cases for each of the ten violated articles, one containing the token `represented' and one without, along with five inadmissible cases. While the article information is available in the task J dataset, we do not use it as it was introduced as a binary violation classification task. 

The rationale annotation process was done using the \textit{GLOSS} annotation tool. The third author of this paper, who is an ECtHR expert, read the case fact statements and highlighted paragraphs which she considered indicative of an eventual finding of a violation for any convention article by the court. Despite our sampling involving randomness, the expert was already familiar with a considerable portion of the decisions. Given this, we abstained from producing a human expert outcome prediction baseline.

\section{Case Sampling for Qualitative Evaluation}
\label{app:case-sampleing}
For the qualitative evaluation of Task A|B, we sample 40 cases (20 each) from validation and test split. In each split, we sample two cases for each of the ten \textit{allegedly violated} articles, one with a finding of a convention violation and with a non-violation finding.

\section{Decision Tree Performance}
\label{app:DT-performance}
\begin{figure*}
        \centering
        \begin{subfigure}[b]{0.475\textwidth}
            \centering
            \includegraphics[width=\textwidth]{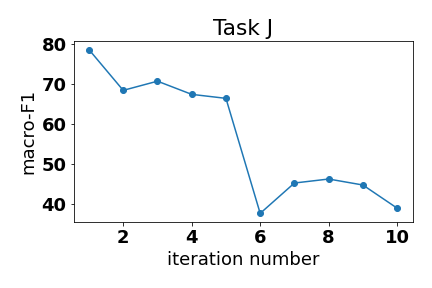}
            \label{fig:mean and std of net14}
        \end{subfigure}
        \hfill
        \begin{subfigure}[b]{0.475\textwidth}  
            \centering 
            \includegraphics[width=\textwidth]{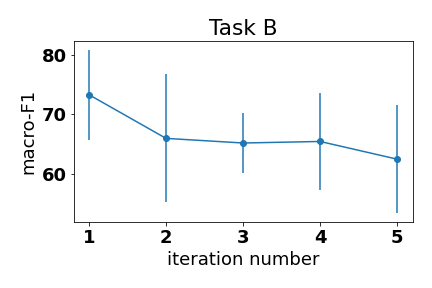}
            \label{fig:mean and std of net24}
        \end{subfigure}
        \vskip\baselineskip
        \begin{subfigure}[b]{0.475\textwidth}   
            \centering 
            \includegraphics[width=\textwidth]{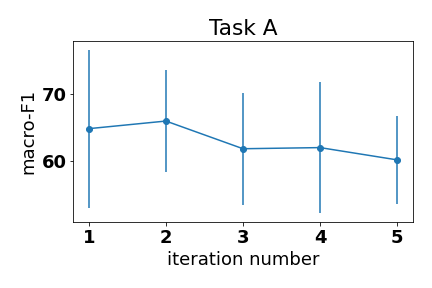}
            \label{fig:mean and std of net34}
        \end{subfigure}
        \hfill
        \begin{subfigure}[b]{0.475\textwidth}   
            \centering 
            \includegraphics[width=\textwidth]{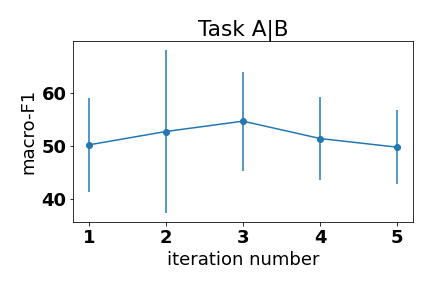}
            \label{fig:mean and std of net44}
        \end{subfigure}
        \caption[]
        {\small Macro-F1-Performance of Decision Trees across different iterations with removal of informative feature nodes in successive iterations. For Task A,B, A|B, standard deviation bars represent variability in F1-score across the 10 articles.} 
        \label{fig:dt}
    \end{figure*}

Figure \ref{fig:dt} shows the  performance of our decision tree model across iterations for different tasks. After each iteration, we remove the informative tokens from previous iterations. In case of task J, we notice a steep fall after iteration 5. Tasks A and B exhibit less dramatic falloff of macro-averaged F1, owing to the different nature of the tasks as article-specific violations. Performance on Task A|B even shows small increases, albeit with a low absolute score. The large standard deviations bands computed across all articles show considerable variation.
    
 \section{Spurious Vocabulary identified by Expert }
\label{app:spurious-vocab}   
Following is the spurious vocabulary we obtained with respect to each task. 
\begin{itemize}
\item\textbf{Task J:}  represented,  national, mr, summarised, practising, lawyer, agent, paragraph

\item \textbf{Task B:}  hearing, born, adjourned, detained, noted, hearing, alleged, investigation, place, question, members

\item \textbf{Task A:} stated, could, also, one, arrested, detained, hearing, investigation, hearing, within, due, second, hearing, certain

\item \textbf{Task A|B:}  february, published, november, march, religious, investigation, first, service, letter, carried, would, one, submitted, head, march, damage, group,provided, seen
\end{itemize}

The words were chosen as relevant or irrelevant by using the daily vocabulary of a human rights lawyer working at the ECtHR as a reference. A word was considered legally relevant if, taken individually, it could be introduced into legal reasoning. For instance, the word ``religious'' was spurious because taken individually it says nothing about the content of a norm. One may talk about religious freedom, but the legally relevant word there is freedom. Article 9 mentions religion, but restrictions related to religion may also be present under Article 8, 3, 2, 5, etc. Under the same Article 9 for instance, the court decides whether there has been a violation depending on criteria such as tolerance, pluralism, etc. It is those criteria that are relevant whereas ``religion'' is not by itself relevant as a part of the legal reasoning.

\section{LMI Calculation}
\label{app:lmi}  
We calculate LMI for each pair of 
token $t$ and label $y$ as follows:
\begin{equation}
    LMI(\text{t, y}) = p(\text{t, y}) \times PMI(\text{t, y})
\end{equation}
\begin{equation}
    p(\text{t, y}) = \frac{\text{count(t,y)}}{|D|} 
\end{equation}
\begin{equation}
    PMI(\text{t, y}) = \log \frac{p(\text{t | y})}{p(\text{t})}
\end{equation}
where $count(t, y)$ denotes the co-occurrence of $t$ and label $y$, and $|D|$ is the number of unique words in the training set.

In the case of binary classification (task J) and one-vs-one multi-label classification (task A|B), we calculate the LMI score for a token as the absolute difference between LMI scores for both positive and negative labels, as both the labels represent a particular class. In one-vs-rest (tasks A, B), we simply take the difference between LMI scores for both positive and negative labels (rather than absolute difference) as the negative label does not specifically represent a particular class. Finally, we calculate the z-score statistic of the effective LMI score for each token to identify significant tokens.

\section{Model configuration \& Training}
\label{app:model-config}  
\noindent {\bf Spurious token identification}: We train a series of decision trees of depth 3 to assemble lists of predictive tokens for expert filtering. The feature vector consists of whitespace-tokenized unigrams reduced by the LMI filtering explained above. For task J, this means training trees that predict the binary violation label. For task A and B we employ a one-vs-rest classification to produce one decision tree series per article. For task A|B we provided the task B labels (allegedly violated articles) in one-vs-one fashion per article, with positive instances being facts where that particular article was deemed violated, and negatives where that particular article was merely alleged but not deemed violated.

\noindent {\bf LJP models}: Our models compute BERT-based word embeddings of size 768. Our word level attention context vector size is 300. The sentence level GRU encoder dimension is 200, thus giving a bidirectional embedding of size 400, and a sentence level attention vector dimension of 200. The final dense classifier for all tasks has 100 hidden units. The output dimension is 1 for task J and 10 for the other tasks (i.e. one per convention article). For task A|B, we concatenate a multi-hot 10-element feature vector containing the task B labels to the output of the feature extractor before it is passed to the classifier. All discriminators (country, length, and vocabulary) are built as analogous classifiers with a hidden dimension of 100 and output layer dimensions as required by each of them. We use mini batches size of 8 in case of Task J and 16 for all other tasks. The model is optimized end-to-end using Adam  \cite{kingma2014adam}. The dropout rate \cite{srivastava2014dropout} in all layers is 0.1. We determine the best learning rate using a grid search on the development set and use early stopping based on the development set F1 score.
   
\section{Visualization of IG score} \label{app:screenshots}
\begin{figure*}[]
\centering

     \begin{subfigure}[b]{\textwidth}
         \centering
         \includegraphics[width =0.9\textwidth]{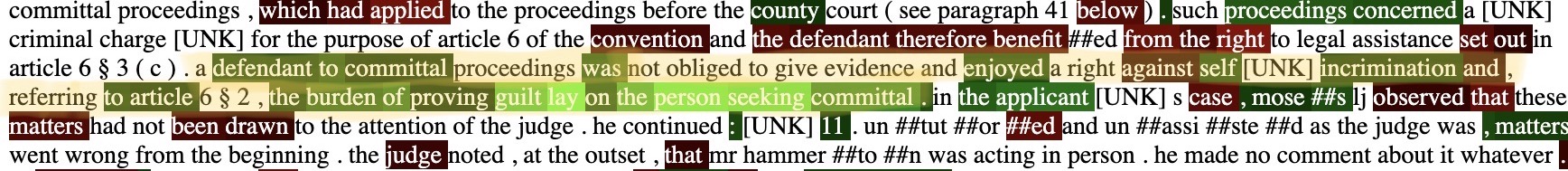}
         \caption{gradAll}
         \label{fig:gradAll}
     \end{subfigure}

     \begin{subfigure}[b]{\textwidth}
         \centering
         \includegraphics[width =0.9\textwidth]{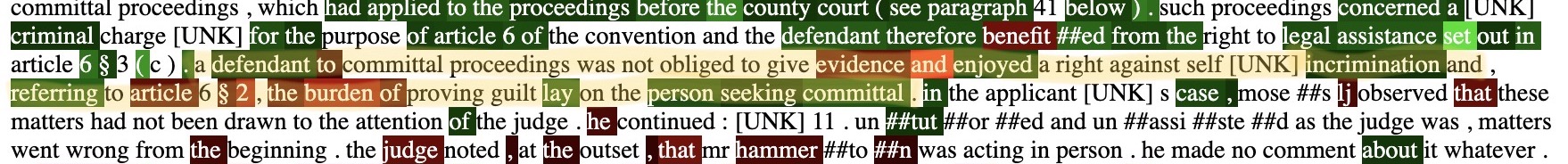}
         \caption{paraRem}
         \label{fig:paraRem}
     \end{subfigure}

{\caption{\label{app:pos} Example visualizations of different IG scores derived from the (a) gradAll and (b) paraRem model, respectively. The gradAll model focuses contiguously and densely on the expert-annotated indicative sentence `A defendant to committal proceedings ... proving guilt lay on the person seeking committal' (yellow background highlighted); while the paraRem model fails to focus the latter half the indicative sentence. }}
\end{figure*}

Figure \ref{app:pos} exhibits screenshot excerpts of a sample case text provided to the legal expert for qualitative evaluation. The yellow background highlight was not in the original visualization and has been supplied here as a reference. We add it here as an example of focus patterns shifting incurred by our deconfounding method.

\begin{table*}[]
    \centering

\begin{tabular}{|c|p{2cm}|p{2cm}|p{2cm}|p{2.5cm}|p{2cm}|}

      \hline
      & & & & & \\
   \textbf{Article} & \textbf{\parbox{3cm}{\% of cases  \\ alleged}} &\textbf{\parbox{3cm}{\% of cases  \\ alleged and  \\ violated}}   &\textbf{\parbox{5cm}{\% of cases \\ alleged but \\ not violated}} &\textbf{\parbox{5cm}{conditional \\ probability \\ of violation \\ given allegation}}  &\textbf{\parbox{5cm}{\% of cases \\ not alleged  \\ but violated}} \\  
    & & & &  & \\
   \hline
10  & 4.9 & 3.1 & 1.8 & 63.27 & 0.13 \\
11  & 1.8 &  1.2  & 0.6 & 66.67 &  0.02 \\
14 &  4.93 & 1.51 & 3.42 & 30.63 & 0.06 \\
2 & 6.92 & 5.48 & 1.44 & 79.13 & 0.13 \\
3 & 19.33 & 14.5  & 4.83 & 75.0 &  0.49 \\
5 & 18.03 & 14.68 & 3.36 &81.39 & 0.52 \\
6 & 60.41 & 51.64 & 8.77 & 85.49 &  0.62 \\
8 & 11.73 & 7.66 & 4.08 & 65.25  & 0.23 \\
9 &  0.9  & 0.44 & 0.46 & 49.38 &  0.01 \\
P1-1 & 17.31 & 15.02 & 2.29 & 86.78 & 0.77 \\
\hline

    \end{tabular}
    \caption{Statistics of LexGlue Train Dataset (Total of 9000 Cases) \cite{chalkidis-etal-2022-lexglue} }
    \label{lexglue-stats}
\end{table*}

\input{img/artefactsInJ}

%% file: img/artefactsInJ.tex
\begin{table*}[]
    \centering

\begin{tabular}{@{}lll@{}}
\toprule
                                     & violation & non-violation \\ \midrule
number of cases                              & 3551      & 3549          \\
avg. number of sentences per case             & 55      & 35\\
number of inadmissable cases                  & 0         & 2608\\          
avg. first paragraph number appeared in the text &6.1 &1.8 \\
percentage of cases containing the word 'represented'  & 0.17     & 0.68    \\ \bottomrule

    \end{tabular}
    \caption{Statistics of violation (label 1) and non-violation (label 0) cases in the training set of Task J. Some inadmissible cases are so short  that no paragraph numbers appear. In such situations we count the paragraph number as 0.}
    \label{artefactsJ}
\end{table*}